\documentclass[runningheads]{llncs}
\usepackage{makeidx}
\usepackage{graphicx}
\usepackage{amsmath,amssymb} 
\usepackage{color}

\usepackage{times}
\usepackage{graphicx} 
\usepackage[tight]{subfigure} 

\usepackage{algorithm}
\usepackage{algorithmic}

\usepackage{hyperref}

\usepackage{epsfig}
\usepackage{amsmath}
\usepackage{amssymb}
\usepackage{mathrsfs}
\usepackage{upgreek}
\usepackage{soul}
\usepackage{rotating}
\usepackage{multirow}
\usepackage{multibib}
\newcites{suppl}{References}
\newcites{corrig}{References}
\usepackage{layouts}
\usepackage{enumitem}
\usepackage{bbding} 
\newcommand{\NEW}[1]{#1}
\DeclareMathSizes{9}{8}{7}{5}
\DeclareMathOperator*{\argmin}{arg\,min}
\newcommand{\N}{ \mathcal{V}}
\newcommand{\E}{ \mathcal{E}}
\newcommand{\SP}{ \mathcal{P}}

\newcommand{\onedot}{.}

\def\eg{\emph{e.g}\onedot} 
\def\ie{\emph{i.e}\onedot}

\begin{document} 
\renewcommand*{\theHsection}{paper.\the\value{section}}
\pagestyle{headings}
\mainmatter

\def\GCPR17SubNumber{43}

\title{Diverse $M$-Best Solutions by Dynamic Programming}


\titlerunning{Diverse $M$-Best Solutions by Dynamic Programming}
\authorrunning{C. Haubold, V. Uhlmann, M. Unser, F. A. Hamprecht}

\author{Carsten Haubold$^1$, Virginie Uhlmann$^2$, Michael Unser$^2$, Fred A. Hamprecht$^1$}
\date{\vspace{-0.3in}}

\institute{$^1$ University of Heidelberg, IWR/HCI, 69115 Heidelberg, Germany.\\
$^2$ \'Ecole Polytechnique F\'ed\'erale de Lausanne (EPFL), BIG, 1015 Lausanne, Switzerland.}

\maketitle

\begin{abstract} 
Many computer vision pipelines involve dynamic programming primitives such as finding a shortest path or the minimum energy solution in a tree-shaped probabilistic graphical model. In such cases, extracting not merely the best, but the set of $M$-best solutions is useful to generate a rich collection of candidate proposals that can be used in downstream processing.
In this work, we show how $M$-best solutions of tree-shaped graphical models can be obtained by dynamic programming on a special graph with $M$ layers. The proposed multi-layer concept is optimal for searching $M$-best solutions, and so flexible that it can also approximate $M$-best diverse solutions.
We illustrate the usefulness with applications to object detection, panorama stitching and centerline extraction.\\

\emph{Note:} We have observed that an assumption in \ref{sec:k1M} is not always fulfilled, see the corrigendum on page \pageref{chap:corrigendum} for details.
\end{abstract}

\section{Introduction}\label{sec:intro}

A large number of problems in image analysis and computer vision involve the search for the \emph{shortest path} (\eg, finding seams and contours) or for the \emph{maximum-a-posteriori} (MAP) configuration in a tree structured graphical model, 
as in hierarchies of segmentation hypotheses or deformable part models. To compute the solution to those problems, 
one relies on efficient and optimal methods from dynamic programming~\cite{Bellman1952} such as Dijkstra's algorithm~\cite{Dijkstra1959}.
In many of these scenarios, it is of interest to find not merely the single lowest energy (\ie, MAP) solution, but the $M$ solutions of lowest energy (\emph{$M$-best})~\cite{Lawler1972,Seroussi1994,Nilsson1998,Rollon2011,Batra2012b}. This can \eg~be useful for learning~\cite{Lampert2011}, tracking-by-detection methods that allow competing hypotheses~\cite{Milan2013,Jug2014,Schiegg2014}, or for re-ranking~\cite{Yadollahpour2013} solutions based on higher order features which would be prohibitively complex for the original optimization problem.
If these $M$ solutions are required to differ in more than one label, the problem is referred to as \emph{diverse $M$-best}~\cite{Batra2012a,Kirillov2015,Prasad2014}.

\paragraph{Contributions:}\label{sec:contributions}

In this work, we show how the optimal second best ($M=2$) solution of a tree-shaped graphical model can be found through dynamic programming in a multi-layer graph by using a replica of the original graph as second layer and connecting both layers through edges with special jump potentials (Section~\ref{sec:k1M2}). Using these building blocks, we extend our approach to exactly find the $M>2$ best solutions sequentially by constructing $M$-layer graphs (Section~\ref{sec:k1M}). 
While the above can be seen as a special case of~\cite{Yanover2004}, our multi-layer approach is an intuitive interpretation that allows flexible modeling of the desired result.
We thus develop two heuristics using multiple layers to find the approximate diverse $M$-best solutions for tree-shaped graphical models (Section~\ref{sec:kM2}). Lastly, we experimentally compare the different diversity approximations to prior work, and show results for a variety of applications, namely: \emph{i)} panorama stitching, \emph{ii)} nested segmentation hypotheses selection, and \emph{iii)} centerline extraction (Section~\ref{sec:app}).

\section{Related Work}\label{sec:relatedWorks}

\paragraph{M-best MAP:} 
An algorithm for sequentially finding the $M$ most probable configurations of general combinatorial problems was first presented in \cite{Lawler1972}. To find the next best solution, they branch on the state of every single variable, resolve, and finally choose the best of all resulting configurations. While this works for any optimization method and model, it is in practice prohibitively expensive.
Several works have extended this to junction trees~\cite{Seroussi1994,Nilsson1998} that work in $O(|\N|(L^2+M+M\log(|\N|M)))$, while~\cite{Rollon2011,Flerova2012} developed a similar bucket elimination scheme ($O(M|\N|L^{|\N|})$). 
A similar idea was applied in~\cite{Eppstein1998} to find the $M$ shortest paths jointly by building an auxiliary graph with a heap at every node that contains the $M$-best paths to reach that node. 
For situations where the optimal or approximative max-marginals can be computed,~\cite{Yanover2004} derived an improvement on~\cite{Nilsson1998} such that the max-marginals have to be computed only $2M$ times, yielding the same runtime complexity ($O(M|\N|L^2)$) as the method we present here. A method that finds the $M$ best solutions on trees in only $O(L^2 V + \log(L)|\N|(M­-1))$ by an algebraic formulation that is similar to sending messages containing $M$ best values as in~\cite{Flerova2012} was presented by \cite[Chapter 8]{schlesinger2013ten}.
However, in contrast to~\cite{Yanover2004,schlesinger2013ten}, our approach provides a lot of modeling flexibility, allowing it to be used to approximate diverse $M$ best solutions as well.
A polyhedral optimization view of the sequential $M$-best MAP problem is given in~\cite{Fromer2009}. There, a linear programming (LP) relaxation is constructed by characterizing the assignment-excluding local polytope through spanning-tree-inequalities.
This LP relaxation is tight for trees for $M=2$, but not for higher $M$ or loopy graphs because the assignment-excluding inequalities could together cut away other integral vertices of the polytope. An efficient message passing algorithm for the same LP relaxation exploiting the structure of the polytope was designed by~\cite{Batra2012b}.
A completely orthogonal way to explore solutions around the optimum would be to sample from the modeled distribution, \eg~using Perturb-and-MAP~\cite{Papandreou2011}.

\paragraph{Diverse $M$-Best:}
For general graphical models, the first formulation of the \emph{diverse $M$-best} problem 
can be found in~\cite{Batra2012a}.
Even though their Lagrangian relaxation of the diversity constraint can work with any choice of metric, 
it is not even tight for Hamming distances of $k>1$.
Different diversity metrics are explored in~\cite{Prasad2014}, where a greedy method to find good instances 
from the (exponentially big) set of possible solutions is designed by setting up a factor graph with higher 
order potentials, assuming that the diversity metric is submodular.
In~\cite{Kirillov2015}, the authors construct a factor graph that \emph{jointly} finds the diverse $M$-best 
solutions by replicating the original model $M$ times and inserting factors for the diversity penalty depending 
on the structure of the chosen distance. 
It is shown that maximizing the diversity of the $M$-best solutions jointly, not only sequentially as in~\cite{Batra2012a}, can yield better results. 
They propose a reformulation that preserves solvability with $\alpha$-$\beta$-\emph{swap}-like methods. Still, when applied to trees, 
the factors introduced for diversity unfortunately turn the problem into a loopy graph and prevent the application of dynamic programming.
All those approaches incorporate the diversity constraint into the original optimization problem by Lagrangian relaxation.
In contrast, the constructions we propose in this work yield a solution with the desired diversity in a single shot.
A different line of work \cite{Chen2013,Chen2014} focused on extracting the $M$-best \emph{modes}, but those methods' computational complexity renders them intractable for large graphs.
\section{Optimal Second Best Tree Solutions}\label{sec:k1M2}

We now present how the second best solution of a tree-shaped graphical model can be obtained using dynamic programming on a special graph construction. By second best, we mean a solution that differs from the best configuration in at least one node, \textit{i.e.}, that has a Hamming distance of $k \geq 1$ to the best solution. 
We begin with an informal motivation based on the search for the second best shortest path.

\paragraph{Motivation:}
\begin{figure}[t]
\centering
\includegraphics[width=0.8\linewidth]{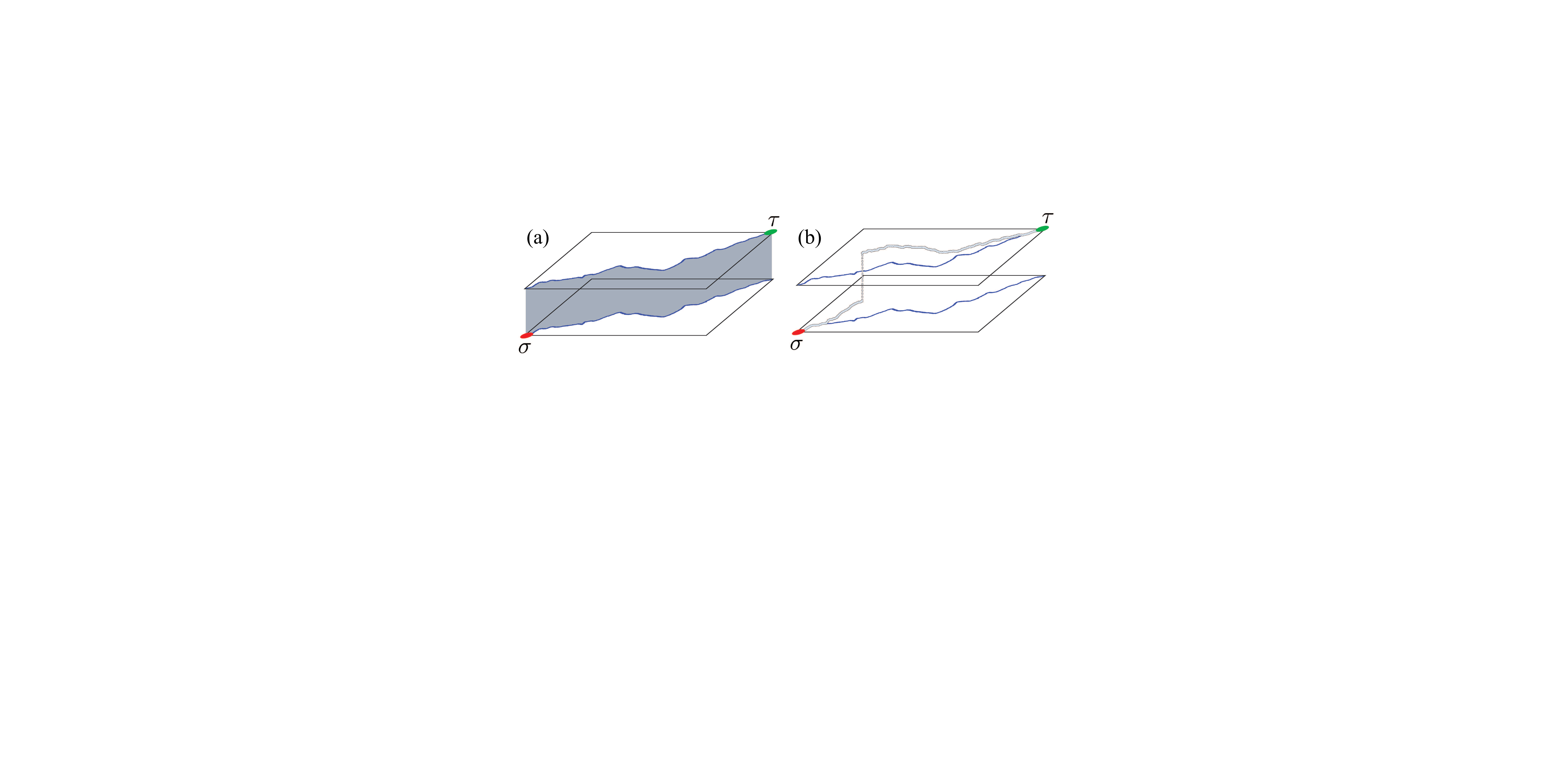}
\caption{
A schematized two-layer grid-graph construction to find the second best shortest path from the source $\sigma$ in the lower, to the target $\tau$ in the upper layer. A valid path is required to jump between layers, which is allowed everywhere for the best path.
\textbf{(a)} Shown in blue is the best solution, which could have jumped to the upper layer at every node along the path with the lowest cost. 
\textbf{(b)} To find the second best path, layer jumps are forbidden at the nodes used by the best solution. Thus the second path diverges to the jump location leading to the next minimal cost path.}
\label{fig:layeredShortestPaths}
\end{figure}

The Dual Dijkstra method from~\cite{Fujita2003} allows finding not only the best, but a collection of $M$ shortest paths from a source $\sigma$ to a target $\tau$ in a graph. To do so, two shortest path trees are constructed, one starting at the source and one at the target. Thus, for every node $v$, the shortest path from the source $\SP_{\sigma,v}$ and to the target $\SP_{v,\tau}$ is known. Summing the distances to source and target gives the length of the shortest path from $\sigma$ to $\tau$ via $v$. An important property is that, for all vertices along the shortest path from $\sigma$ to $\tau$, this sum is equal to the length of the shortest path.

\NEW{Now imagine these shortest path trees as two copies of the initial graph stacked as two layers, as seen in Figure~\ref{fig:layeredShortestPaths}\emph{(a)}. The lower layer indicates the lowest cost to reach every vertex $v$ from $\sigma$, and the upper layer the cost of the shortest path to reach $\tau$ from $v$. By selecting any vertex $v$ and connecting the paths at $v$ in the lower and upper layer, one can again find the shortest path from $\sigma$ to $\tau$ via $v$, this time by introducing an auxiliary \emph{jump} edge between the two layers.}

The benefit of this two layer setup is that to find the \emph{second} best solution, we simply have to search for the vertex that \emph{does not} lie on the best path, at which jumping between the layers leads to the minimal cost path.

\begin{figure}[t]
\centering
\includegraphics[width=0.95\linewidth]{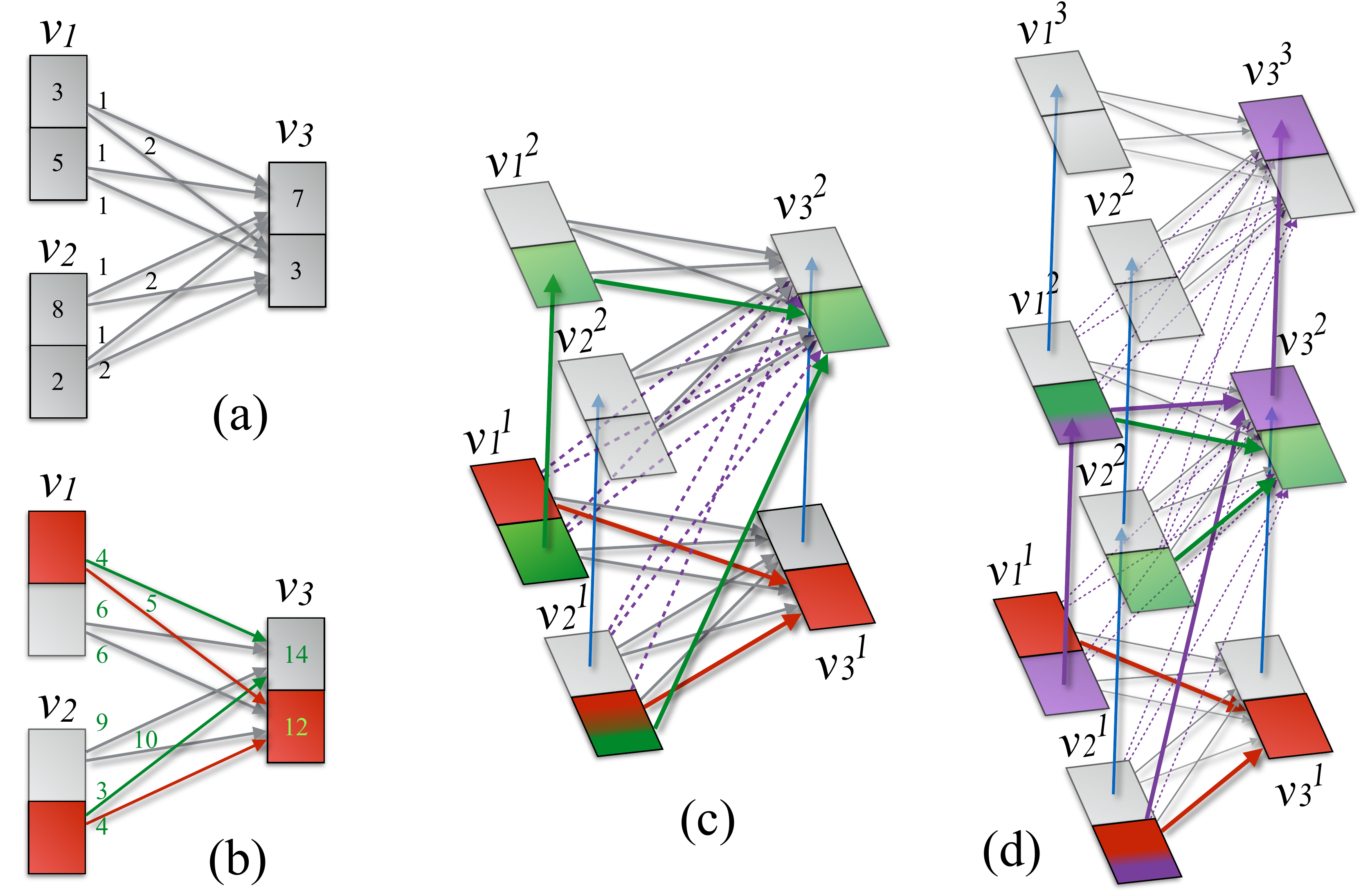}
\caption{\textbf{(a)} Minimal tree structured graph of nodes $v_1$, $v_2$, and $v_3$, with two states each, visualized as stacked boxes. $v_3$ is arbitrarily designated as the root, or target. Unary and pairwise costs are shown as numbers in boxes and along edges, respectively. Its optimal solution is highlighted in red in \textbf{(b)}. Green edges correspond to  the $\argmin$ incoming configurations for each state, and green numbers depict the accumulated min-sum messages. \textbf{(c)} Two-layer tree used to find the second best solution. Blue arcs represent layer-jump-edges with finite potential, which are available at states not occupied by the best solution. Purple dashed edges need to be considered if, at a branching point (such as $v_3^2$), not all incoming messages are coming from the upper layer. The second best solution is represented in green. \textbf{(d)} Searching for the $3^{rd}$-best solution (blue) with a Hamming distance of $k=1$ to the best (red) and second best solution. The new solution must jump twice to reach the upper layer, by taking a state that was not used in the configuration represented by layers 1 and 2.}
\label{fig:two-layer-tree}\label{fig:tree}\label{fig:treeSol1}\label{fig:treeSol2}\label{fig:tree3}
\end{figure}

\paragraph{Dynamic Programming:}\label{sec:generalDP}
Let us briefly review the dynamic programming (DP) paradigm on an undirected tree-shaped graph $\bar{G}=(\N,\bar{\E})$. 
We denote the state of a node $v \in \N$ as $\mathbf{x}_v$, and the full state vector as $\mathbf{x}=\{\mathbf{x}_v: v\in\N\}$.
The potentials of node $v$ (\textit{unary} potential) and of the edge connecting nodes $u$ and $v$ (\textit{pairwise} potential) are represented by $\theta_v(\mathbf{x}_v)$ and $\theta_{uv}(\mathbf{x}_u, \mathbf{x}_v)$, respectively.
From this, we define the inference problem as an energy minimization task~\cite{Koller2009} with objective
\begin{equation}\label{eq:obj}
\min_\mathbf{x} \sum_{v\in\N} \theta_v (\mathbf{x}_v) + \sum_{(u,v)\in\bar{\E}} \theta_{uv}(\mathbf{x}_u,\mathbf{x}_v).
\end{equation}
When applying dynamic programming, one successively computes the energy $E$ of optimal solutions of subproblems of increasing size. One node of the graph $\bar{G}$ is arbitrarily selected as the root node $r$. This results in a directed graph $G=(\N,\E)$ where edges point towards the root. Let $\overleftarrow{N}(v)$ denote the neighboring nodes along incoming edges of $v$ in $G$. Using the tree-imposed ordering of edges, one starts processing at the leaves and sends messages embodying the respective subproblem solutions towards the root. Whenever a node $v$ has received a message from all incoming edges, it can -- disregarding its successors in $G$ -- compute the lowest energies $E_v(\mathbf{x}_v)$ of the subtree rooted at $v$ for every state $\mathbf{x}_v$, and send a message to its parent~\cite{Pearl1988}.
Because leaves have no incoming edges, their energy is equal to their unary potentials. All subsequent nodes combine the incoming messages with their unary potentials to obtain the energy of the subtree rooted at them by
\begin{align}\label{eq:DP}
E_v(\mathbf{x}_v)&:=\theta_v(\mathbf{x}_v)+\sum_{u\in \overleftarrow{N}(v)} \min_{\mathbf{x}_u} \left[ \theta_{uv}(\mathbf{x}_u,\mathbf{x}_v) + E_u(\mathbf{x}_u) \right].
\end{align}

While sending these messages, each node $v$ stores which state ($\argmin_{\mathbf{x}_u}$) of the previous node $(u\in \overleftarrow{N}(v),v)$ along each incoming edge led to the minimal energy of every state $\mathbf{x}_v$.
When the root has been processed, the state that led to the minimal energy is selected and, by backtracking all the recorded $\argmin$, the best global configuration $\mathbf{x}^\star$ can be found. Figure~\ref{fig:treeSol1} shows a minimal tree example.

Regarding DP runtime complexity, consider that \eqref{eq:DP} needs to be evaluated for every state of every node exactly once. In addition, in \eqref{eq:DP}, we consider all states of every incoming edge, of which there are $|\E|=|\N|-1$ in a tree. If $L$ denotes the maximum number of states, one obtains $O(|\N|L^2)$.

\paragraph{Two-layer Model:}

Once the optimal solution is found, we might be interested in the second best solution $\mathbf{x}$, which assigns a different state to at least one node $\exists v\in\N: \mathbf{x}_v \neq \mathbf{x}_v^\star$. 
Because messages in DP only convey the optimal subtree energies, we cannot immediately extract this second best solution. Hence we are looking for a way to enforce that a different state is attained at least once, but we do not know at which node(s) this should happen to yield the optimal energy.
Fortunately, we can apply the same idea as in the second shortest path example: We duplicate the graph to get a second layer and insert edges connecting the two layers such that jumping is only permitted at states not used in the optimal solution $\mathbf{x}^\star$. After propagating messages through both layers, the second best solution can be obtained by backtracking from the minimum energy state of the root in the second layer to leaves in the first layer. This means that messages must have jumped to the second layer at least once at some node $v$ with a state different to $\mathbf{x}_v^\star$, fulfilling our requirement for the second best solution.

We here state the layer setup conceptually and provide the formal construction in the Supplementary.
To create the two layers, we duplicate graph $G$ (Figure~\ref{fig:two-layer-tree}\textbf{a}) such that we get a layer $1$, and a layer $2$ replica. We address the instances of every node $v\in\N$ by $v^1$ and $v^2$ for layer $1$ and layer $2$, respectively. When duplicating the graph, the unary and pairwise potentials of nodes and edges are copied to layer $2$. At every node $v\in\N$, we insert a \emph{layer-jump-edge} from $v^1$ to $v^2$ (blue edges in Figure~\ref{fig:two-layer-tree}\textbf{c}) with a pairwise potential $\theta_{v^1v^2}$ that is only zero if both variables take the same state $\mathbf{x}_{v^1}=\mathbf{x}_{v^2}$ different from $v$'s state in $\mathbf{x}^\star$, and infinity (forbidden) otherwise. This way, finite valued messages in layer two represent configurations that did differ from $\mathbf{x}^\star$ at least once.
These jump edges would suffice for a chain graph, but the branching points in a tree need special consideration. When a layer $2$ branching point is not reached by a layer jump, the current construction only allows considering incoming messages from layer $2$. However, since we only require \emph{one} variable to take a new state, only one branch is necessary to reach layer $2$ on a path with finite cost. To cope with this situation, we insert \emph{layer-crossing} edges from $u^1$ to $v^2$ for all edges $(u,v)\in\E$ (dashed purple edges in Figure~\ref{fig:two-layer-tree}\textbf{c}) with the same pairwise potential as in the original graph $\theta_{u^1v^2}=\theta_{uv}$, and alter the DP update equation for nodes in layer $2$ to
\begin{align}
E_{v^2}(\mathbf{x}_{v^2})&:=\min\bigg(\theta_{v^1v^2}(\mathbf{x}_{v^1},\mathbf{x}_{v^2})+E_{v^1}(\mathbf{x}_{v^1}), \label{eq:E2jump} \\
&\theta_{v^2}(\mathbf{x}_{v^2})+\min_{\substack{L_2 \subseteq \overleftarrow{N}(v)\\ |L_2| \geq 1}}\sum_{u\in L_2}\min_{\mathbf{x}_{u^2}}\left[ \theta_{u^2v^2}(\mathbf{x}_{u^2},\mathbf{x}_{v^2})+E_{u^2}(\mathbf{x}_{u^2}) \right] \nonumber\\
&+ \sum_{u\in \overleftarrow{N}(v) \setminus L_2}\min_{\mathbf{x}_{u^1}}\left[ \theta_{u^1v^2}(\mathbf{x}_{u^1},\mathbf{x}_{v^2})+E_{u^1}(\mathbf{x}_{u^1}) \right] \bigg).\label{eq:Enu2}
\end{align}
Compared to \eqref{eq:DP}, we now have two options instead of one at every node $v$ in layer $2$. Firstly, we can reach $v^2$ by a layer jump. Note that, in case of a jump \eqref{eq:E2jump}, we do not account for the unary $\theta_{v^2}(\mathbf{x}_{v^2})$ as $E_{v^1}(\mathbf{x}_{v^1})$ contains the same term already. Alternatively, at least one of the incoming messages must come from a nonenpty set $L_2$ of predecessors in layer $2$ \eqref{eq:Enu2}, while the remaining messages could \emph{cross} layers.
These options are visualized in Figure~\ref{fig:two-layer-tree}\textbf{c}.

\paragraph{Optimality and Runtime:}
By duplicating the directed graph and inserting two sets of new edges which are oriented towards the root in layer two, the topology of the graph remains a directed acyclic graph, and DP hence yields the optimal configuration. As long as the solution has finite energy, no forbidden \emph{layer-jump-edge} is used, giving us the second best solution. In terms of runtime complexity, we have duplicated the number of vertices and have four times as many edges, which are small constant factors that disappear in $O(|\N|L^2)$. For optimal performance, one can reuse the messages in layer 1 because these do not change.

\section{Optimal $M$-Best Tree Solutions}\label{sec:k1M}
The two-layer setup can easily be extended to multiple layers, which allows us to search for the $M$-best solutions with a Hamming distance of $k \geq 1$.
We use one additional layer per previous best configuration; that is, $M$ layers. Each layer is responsible for one of the previous solutions, hence its \emph{layer-jump-edges} are restricted according to the respective solution. Solutions must be ordered by increasing cost; such that the first layer constrains jumps with respect to the best configuration, the second layer for the second best, and so on. The new update rules from in Section \ref{sec:k1M2} can then be applied to every consecutive pair of layers. Figure~\ref{fig:tree3}\textbf{d} shows an example.

\paragraph{Optimality and Runtime:} 
When considering more than one previous solution using the multi-layer setup, the jump restrictions encoded in the \emph{layer-jump-edge} potentials are independent at each layer. For any given node and state in a layer, the cost and path to reach it are optimal with respect to all layers below. This straightforwardly holds for the one and two layer cases, and is the reason why layers must be ordered by increasing cost of the represented previous solutions. For the sake of argument, layers could be flattened as they are getting processed, bringing back the problem to a series of $M-1$ optimal two-layer cases, which yields a computational complexity of $O(M|\N|L^2)$.

\section{Approximate Diverse $M$-Best Solutions}\label{sec:kM2}
In the classical diverse-$M$-best setting~\cite{Batra2012a}, additional solutions are required to have~\eg~a Hamming distance of $k>1$. Here, we look at the straightforward multi-layer extension of Section~\ref{sec:k1M2} to handle $k>1$. We argue that this approach is suboptimal, and present a two-layer approximation that trades quality for efficiency. Lastly, we discuss how this could be used to find $M$ diverse solutions.

\paragraph{Multi-layer Model:}
To ensure that the next solution differs by at least $k$ from the best one, we could construct a $k+1$-layer graph using the same jumping criteria between all layers. To reach the top layer, a solution must hence jump $k$ times. This raises two challenges: \emph{(a)} a branching point at layer $N$ can be reached by a combination of edges from different layers such that the predecessors \emph{in total} account for a Hamming distance of $N$, and \emph{(b)} a solution should never jump more than once at a single node, otherwise it will not have the desired diversity.
Both can be achieved by adjusting the DP update equation to consider a set of admissible incoming edge combinations. We provide the precise expression in the Supplementary. 

Unfortunately, this simple setup does not yield optimal solutions. To forbid two jumps in a row, one needs to introduce a dependence on a previously made decision. These dependencies invalidate the subproblem optimality criterion for DP to yield the correct result. It is thus possible that DP does not reach the root on layer $k$ with finite cost, as shown in the Supplementary. Using the same reasoning, even if a valid solution is found, it is not necessarily optimal.
Additionally, the set of admissible combinations of incoming edges grows combinatorially, making this approach unsuited for large $k$.

\paragraph{Diversity Accumulation:}
Instead of using $k$ layers, one can also formulate a heuristic on a two-layer graph that ensures that any found solution contains the desired amount of diversity.
To do so, we reformulate the Hamming distance constraint (that the new solution must differ from the previous one at $k$ nodes) as a constraint on accumulated diversity, \ie, that $\sum_{v \in \N} \alpha_v(\mathbf{x}_v)  > T$, where $\alpha$ is a measure of diversity per node and state, and $T$ a threshold.
We change the DP update rules as follows. First, while propagating messages from the leaves of the tree to the root in layer $1$, one must also propagate the amount of diversity accumulated by the corresponding configuration of the subtree. Let us denote nodes and edges of the subtree rooted at node $v$ in layer $1$ by $\overleftarrow{\N_{v^1}}$ and $\overleftarrow{\E_{v^1}}$, respectively. The accumulated diversity $\mathcal{A}$ is given by 
\begin{equation}
\mathcal{A}_{v^1}(\mathbf{x}_{v^1}):=\sum\limits_{i\in\overleftarrow{\N_{i^1}}} \alpha_{i}(\mathbf{x}_i) + \sum\limits_{(i,j)\in\overleftarrow{\E_{i^1}}} \alpha_{ij}(\mathbf{x}_i, \mathbf{x}_j).
\end{equation}
Then, we define the layer jump potential $\tilde{\theta}_{v^1v^2}(\mathbf{x}_{v^1},\mathbf{x}_{v^2})$ to be infinity as long as the accumulated diversity is below the desired threshold $\mathcal{A}_{v^1}(\mathbf{x}_{v^1}) < k$.
The limitation of this heuristic is that, at each node and state, we find the optimal subtree configuration by minimizing the energy without considering diversity.
This can prevent us from finding solutions with large diversity.
Yet, as we will see in the experiments, this approach has an attractive runtime because it only requires two layers to find a solution with any Hamming distance $k$, and thus has the constant runtime complexity of $O(|\N|L^2)$ per solution.

\paragraph{Extension to $M$ Diverse Solutions:}
Finding $M$ solutions with a Hamming distance of $k$ could be achieved by stacking $M\times(k+1)$ layers, but then the long range dependency problems depicted above are even more prominent.
With diversity accumulation on the other hand, $M$ diverse solutions can be obtained heuristically by using one diversity map $\alpha$ and one accumulator $\mathcal{A}$ per previous solution. The jump criterion must then ensure that enough diversity has been accumulated with respect to each previous solution.

\section{Applications and Experiments}\label{sec:app}
We now evaluate the performance of our heuristics to obtain diverse solutions with prior work, and demonstrate its applicability to several problems in Computer Vision\footnote{See the Supplementary for an application to depth estimation from stereo.}.

\begin{figure}[t]
\begin{center}
   \includegraphics[width=0.95\linewidth]{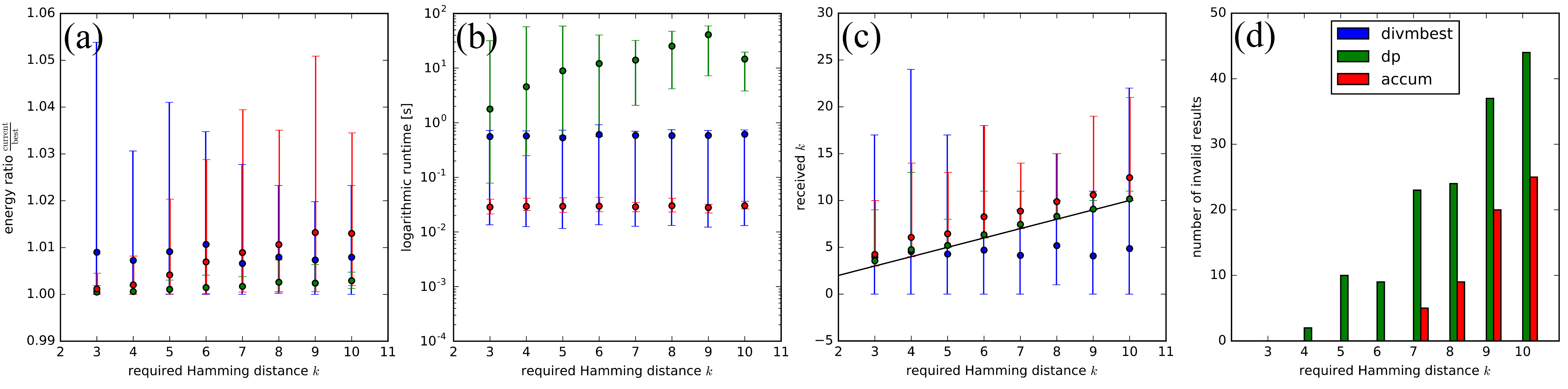}
\end{center}
   \caption{
       Comparison of the $k+1$-layer \texttt{dp} and diversity \texttt{accumulation} heuristic for obtaining diverse solutions (Section \ref{sec:kM2}) against \texttt{divmbest}~\cite{Batra2012a}. All results show mean, minimum and maximum over the valid solutions obtained for every setting on 50 random trees, where \textbf{(d)} shows the number of experiments that did not find a valid solution. \textbf{(a)} Energy ratio between the optimal unconstrained solution and the one with Hamming distance $k$. \textbf{(b)} Runtime. \textbf{(c)} Hamming distance of the resulting solution. Lower is better in all plots but \textbf{(c)}, where the returned Hamming distance should be close to, or preferably above the drawn diagonal.
   }
\label{fig:randomTreeDiversity}
\end{figure}

\paragraph{Comparison with Existing Works:}
\cite{Batra2012a,Yadollahpour2013,Kirillov2015} search for the diverse-$M$-best solutions by incorporating the diversity constraint via Lagrangian relaxation. Our heuristics follow a different approach and turn the constraint into a lower bound instead of relaxing it. The resulting advantage is that we guarantee the set of solutions to be as diverse as required, at the possible expense of a higher cost or the inability to find a solution at all.
On $50$ random trees, with $100$ nodes each, all nodes having $3$ states with unary and pairwise potentials drawn uniformly from the range $[0,1]$, we evaluate different Hamming distances in Figure~\ref{fig:randomTreeDiversity}. We let the method of \cite{Batra2012a} run for $100$ iterations with a step size of $1/n$ in iteration $n$ or stop at convergence.
In terms of runtime, diversity accumulation stands out as it constantly requires only two layers. Because the distance to the best configuration is not enforced by hard constraints, solutions found by \cite{Batra2012a} often contain too little diversity, yielding a too low mean Hamming distance. Diversity accumulation gives solutions with more diversity than required, and hence also deviates more from the optimal energy. In terms of returned diversity and energy, the multi-layer dynamic programming solution yields favorable results compared to the other two methods, but is unfortunately slower -- it suffers from the combinatorial explosion of admissible edge sets to consider -- and fails to find a valid solution on several trees due to the limitations described in Section \ref{sec:kM2}.

\begin{figure}[t]
\centering
\subfigure[]{\includegraphics[width=.32\linewidth]{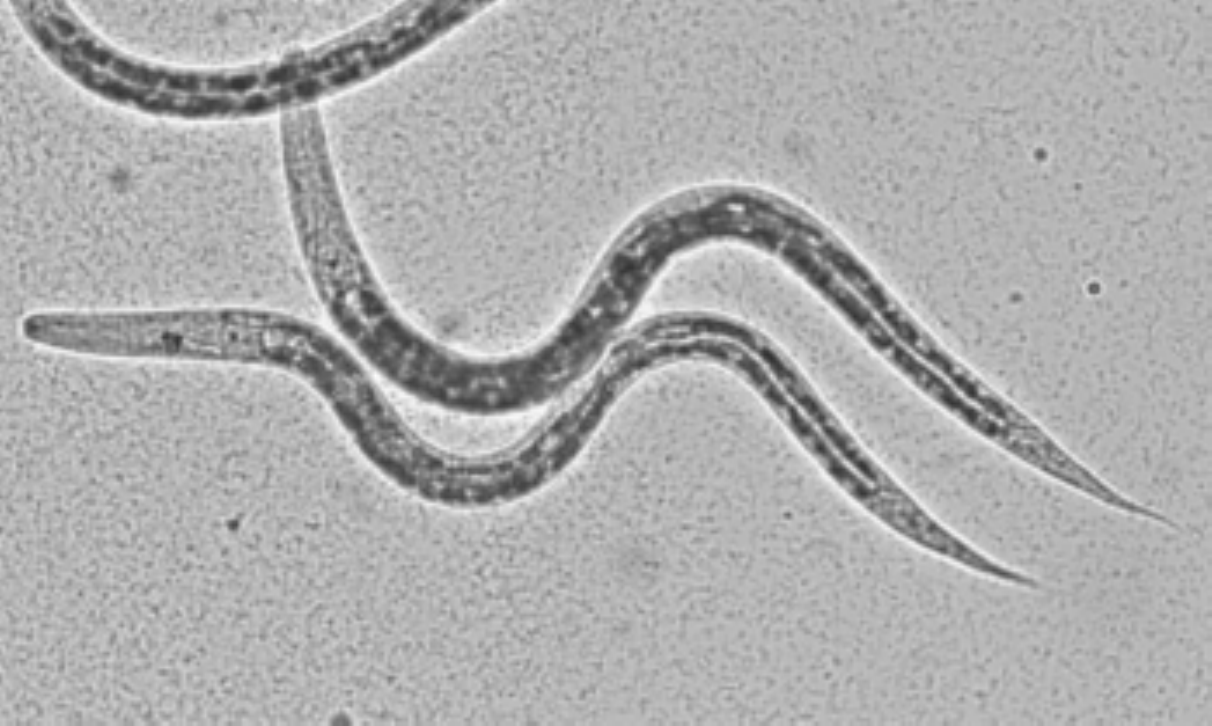}\label{fig:wormRaw}}
\subfigure[]{\includegraphics[width=.32\linewidth]{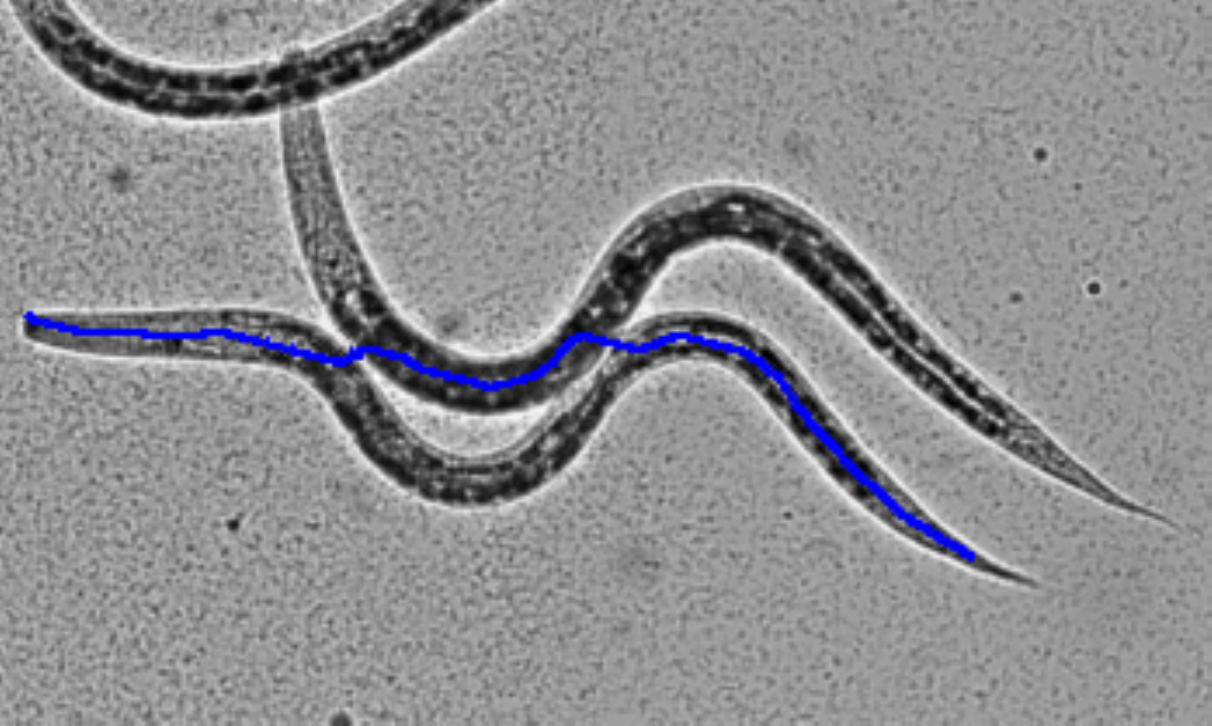}\label{fig:wormSol1}}
\subfigure[]{\includegraphics[width=.32\linewidth]{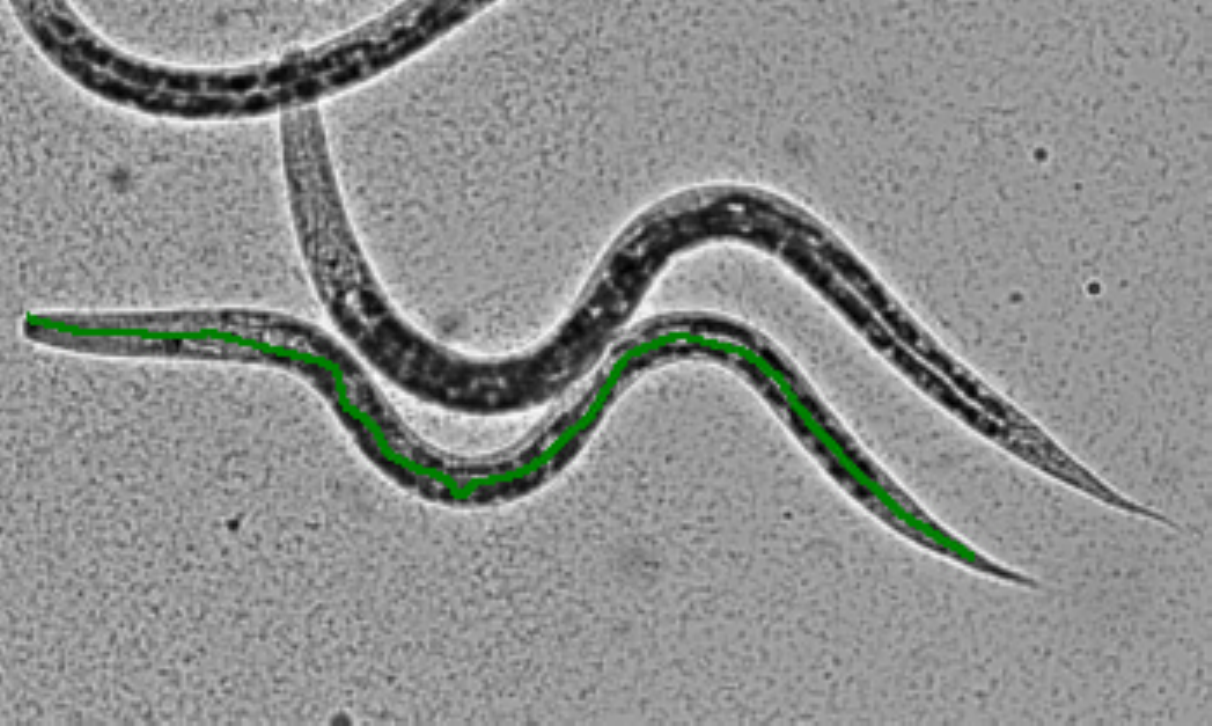}\label{fig:wormSol16}}

   \caption{\label{fig:bioim}Diverse shortest path finding in noisy bioimages featuring objects in close contact.
   \textbf{(a)} Raw brightfield microscope images of \textit{C. elegans}, \textbf{(b)} first best path, and \textbf{(c)} $5$th best path between auto-selected end points using an exclusion corridor of $30$ pixels and a required accumulated diversity of $k\geq~25$.
   }
\end{figure}

%
\paragraph{Medial Axis Identification in Biological Objects:}
Identifying the medial axis of biological objects is a common problem in bioimage analysis, as it serves as a basis for length or growth estimation and tracking-by-assignment. Simple dynamic programming can achieve this task given the end points, although, as biological images tend to get noisy or crowded, designing a robust cost function is difficult. In Figure~\ref{fig:bioim}, we illustrate the usefulness of searching for a collection of possible best solutions instead of only one shortest path in brightfield microscopy images of \textit{C. elegans} nematodes.

\begin{figure}[t]
\begin{center}
   \includegraphics[width=0.8\linewidth]{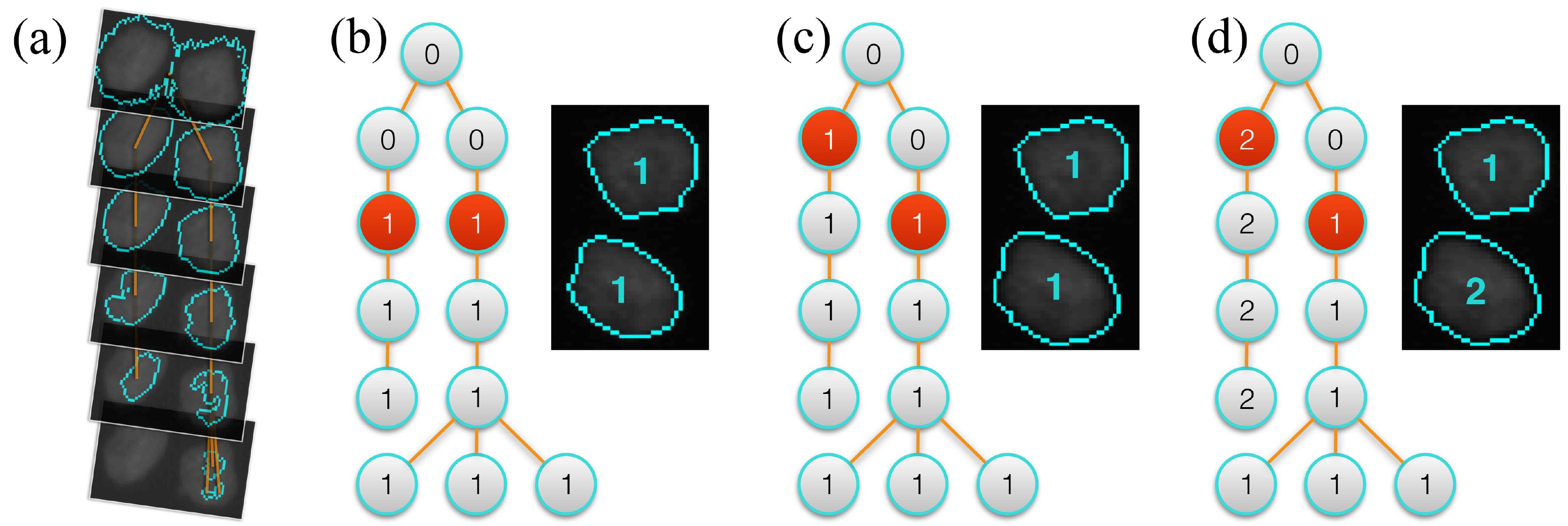}
\end{center}
   \caption{
      Finding the $M$-best configurations of a tree \textbf{(a)} of MSER segmentation hypotheses as in~\cite{Arteta2013}. 
       The best \textbf{(b)}, second \textbf{(c)} and third \textbf{(d)} best configuration found by blocking previous solutions in the respective \emph{layer-jump-edge} potentials. The selected label at each node denotes the predicted object count of the first nonzero ancestor in the tree.
   }
\label{fig:artera}
\end{figure}

\paragraph{Selection of Segmentation Hypotheses:}
In datasets with cell clumps, it is often hard to select the correct detections from a set of segmentation hypotheses. We illustrate this problem in images from the Mitocheck project dataset\footnote{http://www.mitocheck.org/} \cite{Held2010} using the tree model proposed in~\cite{Arteta2013}. There, the task is to assign a class label to each element of a set of nested maximally stable extremal regions. The labels indicate the number of objects that each particular region represents. In the tree, nodes correspond to regions, and edges between parent and child node model the nestedness properties. In Figure~\ref{fig:artera}, we show results obtained when constraining dynamic programming with our $M$-best approach. This is useful to generate segmentation or pose candidates as needed by joint segmentation and tracking procedures, \eg~\cite{Jug2014,Schiegg2014}.

\begin{figure}[t]
\centering

\includegraphics[width=0.8\linewidth]{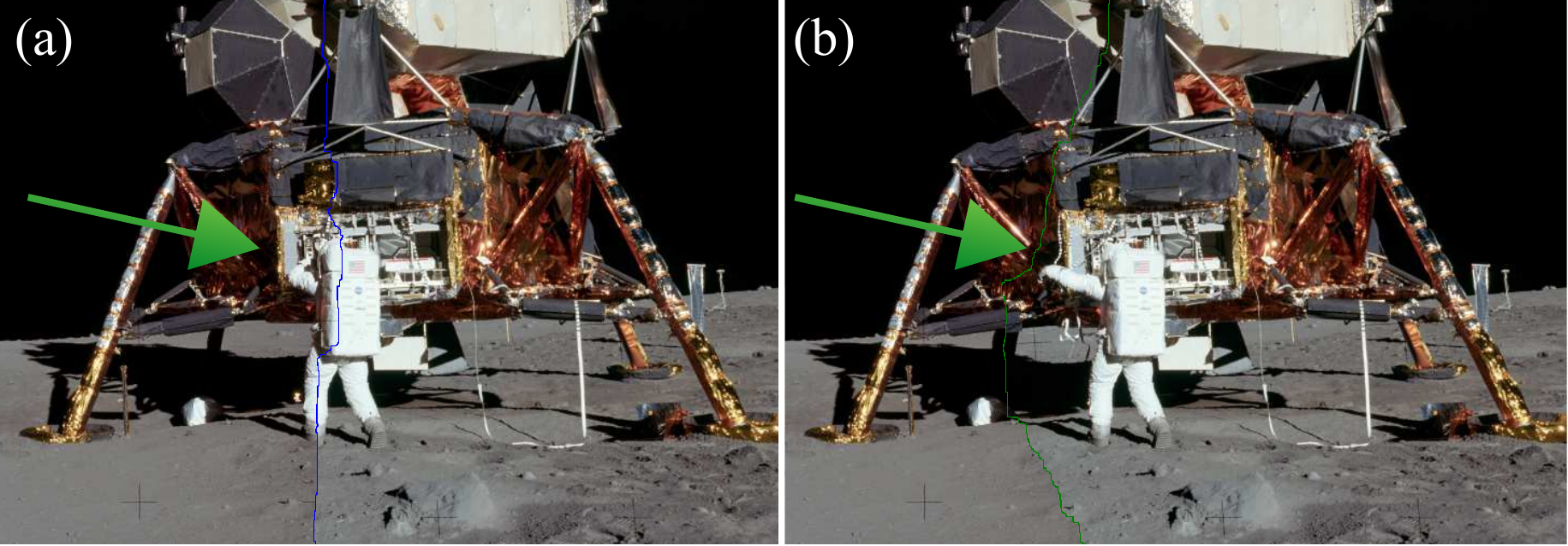}  
\caption{Finding diverse best paths (seams) for panorama stitching. Once the best solution \textbf{(a)} has been found, layer-jump-edges were blocked 
in a corridor around it to obtain the diverse second best solution \textbf{(b)}.
} \label{fig:panorama}
\end{figure}

\paragraph{Panorama Stitching:}
In our motivation in Section~\ref{sec:k1M2}, we mentioned that the proposed multi-layer setup can also be used for shortest paths. Here, we apply that in the context of boundary seam computation for panorama stitching~\cite{Summa2012}. We stitch images taken during the Apollo $11$ moon landing (Apollo-Armstrong: $2$ images of $2349 \times 2366$, courtesy of NASA). As observed in Figure~\ref{fig:panorama}, the second diverse shortest path also corresponds to a visually correct stitching, although the resulting path significantly differs from the globally optimal one.

\section{Conclusion}\label{sec:conclusions}
We have presented a multi-layer graph construction that allows formulating the $M$-best problem for tree-shaped graphical models efficiently through dynamic programming. This flexible framework can be used to find $M$-best solutions for a Hamming distance of $k=1$ optimally. For $k>1$, we present two heuristics, one using a multi-layer graph, and one using two-layers where each new configuration must accumulate diversity before it can reach the upper layer. We evaluated both heuristics against diverse-$M$-best~\cite{Batra2012a}, revealing that both perform favourably with certain strengths over the baseline.
We demonstrated for several practical applications that the presented methods can reveal interesting alternative solutions.

\section*{Acknowledgements} 
This work was partially supported by the HGS MathComp Graduate School, DFG grant HA 4364/9-1,
SFB 1129 for integrative analysis of pathogen replication and spread,
and the Swiss National Science Foundation under Grant 200020\_162343 / 1.

\bibliographystyle{splncs03}
\bibliography{biblio}

\pagebreak

\setcounter{equation}{0}
\setcounter{figure}{0}
\setcounter{table}{0}
\setcounter{section}{0}
\renewcommand*{\theHsection}{suppl.\the\value{section}}
\renewcommand*{\theHfigure}{suppl.\the\value{figure}}
\makeatletter
\renewcommand{\theequation}{S\arabic{equation}}
\renewcommand{\thefigure}{S\arabic{figure}}


\title{Diverse $M$-Best Solutions by Dynamic Programming\\Supplementary}


\author{Carsten Haubold$^1$, Virginie Uhlmann$^2$, Michael Unser$^2$, Fred A. Hamprecht$^1$
}

\institute{$^1$ University of Heidelberg, IWR/HCI, 69115 Heidelberg, Germany.\\
$^2$ \'Ecole Polytechnique F\'ed\'erale de Lausanne (EPFL), BIG, 1015 Lausanne, Switzerland.}
\date{\vspace{-0.3in}}
\maketitle

\section{Optimal Second Best Tree Solutions: Graph Construction}\label{sec:suppl:k1M2}

We formally construct the auxiliary directed graph $\tilde{G}=(\tilde{\N},\tilde{\E})$ composed of two instances of the original directed graph $G=(\N,\E)$ stacked up vertically. We address the lower layer with index $1$, and the upper one with index $2$. The new set of nodes and edges is given by 
\begin{align*}
\tilde{\N}&:=\{ 1, 2 \} \times \N ,\\ 
\tilde{\E}&:=\E^1_{\mathrm{in}} \cup \E^2_{\mathrm{in}} \cup \E_{\mathrm{jump}} \cup \E_{\mathrm{cross}}.
\end{align*}
In the following, the first and second layer copies of an original node $v\in\N$ are written as $v^1$ and $v^2$, respectively.

\begin{eqnarray}
\E^1_{\mathrm{in}}&:=&\{(u^1,v^1)| (u, v) \in \E\} \label{eq:dupE1}\\
\E^2_{\mathrm{in}}&:=&\{(u^2,v^2)| (u, v) \in \E\} \label{eq:dupE2}\\
\E_{\mathrm{jump}}&:=&\{(v^1, v^2) | v \in \N\} \label{eq:layerJump}\\
\E_{\mathrm{cross}}&:=&\{ (u^1, v^2) | (u, v) \in \E\} \label{eq:layerCrossing}\end{eqnarray}

Edges are duplicated for each layer~\eqref{eq:dupE1},~\eqref{eq:dupE2}.
Additionally, we introduce \emph{layer-jump-edges}~\eqref{eq:layerJump} that directly go from any node $v^1$ in layer $1$ to its duplicate $v^2$ in layer $2$.
Lastly, we add edge duplicates that originate in layer $1$ and go to layer $2$. These \emph{layer-crossing-edges} \eqref{eq:layerCrossing} are needed for message passing at branching points.
Unary and pairwise potentials $\tilde{\theta}$ are also replicated for all nodes $v\in\N$ and edges $(u,v)\in\E$ as
\begin{eqnarray*}
\forall v \in \N &&\tilde{\theta}_{v^1}:=\theta_v,\, \tilde{\theta}_{v^2}:=\theta_v \\ 
\forall u,v \in \E &&\tilde{\theta}_{u^1v^1}:=\theta_{uv},\, \tilde{\theta}_{u^2v^2}:=\theta_{uv}, \tilde{\theta}_{u^1v^2}:=\theta_{uv}. 
\end{eqnarray*}

To run dynamic programming on this new graph $\tilde{G}$, messages are propagated starting at all leaves in layer $1$, towards the designated root $r^2\in\tilde{\N}$ on the upper layer. From every node $v^1$ in the lower layer, a message -- embodying the partial solution of the subtree rooted at $v^1$ in layer $1$ -- is propagated in three directions: directly to its successors within layer $1$, crossing layers to the successors' duplicates in the upper layer, and as a jump to this node's duplicate $v^2$ subject to a user-specified jumping criterion.

In summary, in the lower layer, standard messages are being sent as in the lowest energy solution. In the upper layer, every junction point is reached by three kinds of messages: those from within layer $2$, those that are incoming from predecessors in layer 1, and those along layer jump edges between node duplicates. 
At junction points in layer $2$, these incoming messages from both layers must be combined. The important requirement for a valid configuration of any layer $2$ junction point is that \emph{at least one} of the incoming messages must have come from layer $2$ (we denote this \emph{nonempty} set of predecessor nodes as $L_2$), and must thus have \emph{jumped} to layer $2$ in the subtree rooted at this junction point. 
We therefore change the DP rules in layer $2$ to
\begin{align}
E_{v^2}(\mathbf{x}_{v^2})&:=\min\bigg(\tilde{\theta}_{v^1v^2}(\mathbf{x}_{v^1},\mathbf{x}_{v^2})+E_{v^1}(\mathbf{x}_{v^1}), \label{eq:E2jump} \\
&\tilde{\theta}_{v^2}(\mathbf{x}_{v^2})+\min_{\substack{L_2 \subseteq \overleftarrow{N}(v)\\ |L_2| \geq 1}}\sum_{u\in L_2}\min_{\mathbf{x}_{u^2}}\left[ \tilde{\theta}_{u^2v^2}(\mathbf{x}_{u^2},\mathbf{x}_{v^2})+E_{u^2}(\mathbf{x}_{u^2}) \right] \nonumber\\
&+ \sum_{u\in \overleftarrow{N}(v) \setminus L_2}\min_{\mathbf{x}_{u^1}}\left[ \tilde{\theta}_{u^1v^2}(\mathbf{x}_{u^1},\mathbf{x}_{v^2})+E_{u^1}(\mathbf{x}_{u^1}) \right] \bigg).\label{eq:Enu2}
\end{align}
Compared to the standard dynamic programming rules, we now have two options instead of one in layer $2$. Firstly, we can reach the node by a layer jump. Note that, in case of a jump \eqref{eq:E2jump}, we do not account for $\tilde{\theta}_{v^2}(\mathbf{x}_{v^2})$ as $E_{v^1}(\mathbf{x}_{v^1})$ already contains this term. Alternatively, at least one of the incoming messages is coming from layer $2$ using edges from $\E^2_{\mathrm{in}}$ \eqref{eq:Enu2}, while the remaining messages may cross layers and originate from $\E_{\mathrm{cross}}$.

If we choose to set all \NEW{layer-jump-edge} potentials to zero, every vertex qualifies as jump location and we obtain at the root the same solution we would get in the original graph $G$.
If we want to find the second best solution with a Hamming distance of $1$, we set the jump potentials of all states used by the previous solution to infinity. 
Then, a jump can only happen when the current solution differs from the previous one at this node. 
In addition, note that the duplicates of all leaf nodes $v^2$ must be reached via a layer jump. This can be achieved by setting their unaries to infinity $\tilde{\theta}_{v^2}:=\infty$.


\section{Approximate Diverse $M$-Best Solutions}
\begin{figure}[t]
\centering
\includegraphics[width=.4\linewidth]{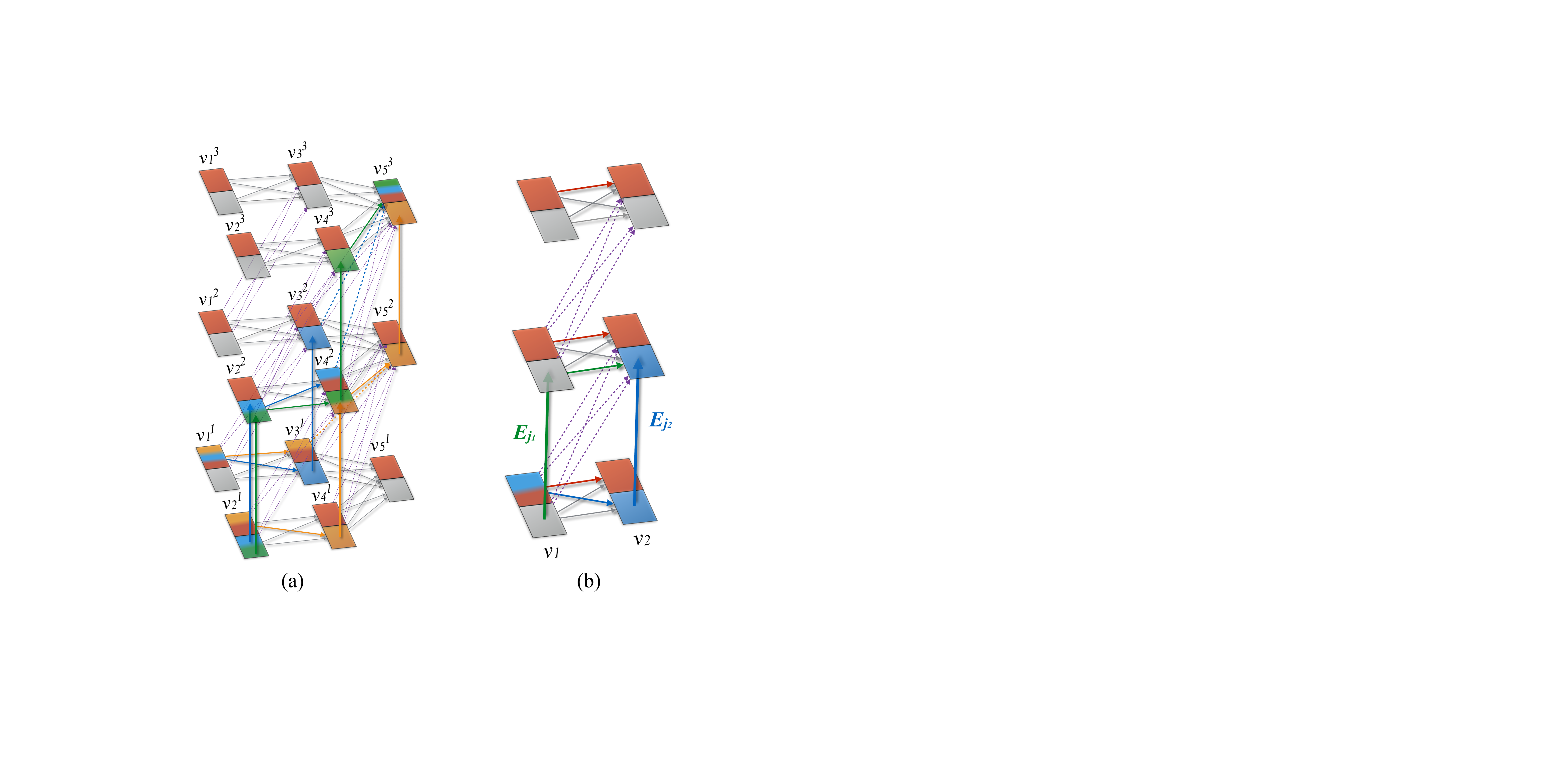}
\caption{\textbf{(a)} Visualization of different ways of obtaining a Hamming distance of $2$, where the red states show the previous solution. To reach
$v_5^3$, one can jump two times at different nodes (green) and arrive in layer $3$, jump once before to layer $2$ and then to $3$ at $v_5$ (orange), or combine two incoming branches from layer $2$ to get
to layer $3$ (blue). \textbf{(b)} Counterexample for $k=2$. If $E_{j_2}<E_{j_1}$, the minimization will pick the predecessors shown in blue, which prevents the algorithm from jumping to layer 3 and finding a valid solution.}
\label{fig:suppl:counterex}
\end{figure}

In \eqref{eq:E2jump} and \eqref{eq:Enu2}, we already considered edges from layer $1$ and $2$ such that at least one of them came from layer $2$ if that specific node was not used for a jump. In the case of $k+1$ layers, we define the set of admissible incoming edge combinations $\mathscr{A}$.
We thus generalize the update equation as follows:
\begin{align}
E_{v^N}(\mathbf{x}_{v^N}):=&\min\bigg(\tilde{\theta}_{v^{N-1}v^N}(\mathbf{x}_{v^{N-1}},\mathbf{x}_{v^N})+E_{v^{N-1}}(\mathbf{x}_{v^{N-1}}) \notag \\ 
&\hphantom{} + \infty\cdot\delta[\mathrm{Pred}_{v^{N-1}}(\mathbf{x}_{v^{N-1}}) == (v^{N-2},\mathbf{x}_{v^{N-2}})], \notag \\
&\tilde{\theta}_{v^N}(\mathbf{x}_{v^N})+\min_{\mathcal{A}\in \mathscr{A}_{v^N}}\min_{\mathbf{x}_a}\sum_{a\in \mathcal{A}} \tilde{\theta}_{av^N}(\mathbf{x}_{a},\mathbf{x}_{v^N})+E_{a}(\mathbf{x}_a) \bigg) \label{eq:UpdateNlayers} \\
\mathscr{A}_{n^N}&=\Bigg\{ \bigg( u_1^{l_1}, u_2^{l_2}, ..., u^{l_{|\overleftarrow{N}(v)|}}_{|\overleftarrow{N}(v)|} \bigg) \, \bigg| \, u_i^{l_i} \in \overleftarrow{N}(v), \, \notag \\ 
&\hphantom{=\Bigg\{ \bigg(}l_i \in \{ 1,...,N \}, \sum_{i=1}^{|\overleftarrow{N}(v)|} (l_i-1)\geq N-1\Bigg\} \label{eq:setA}
\end{align}
To prevent two successive jumps at the same variable \emph{(a)}, one must incorporate a check whether a node was reached by a jump. We thus include a dependence on the previous step. We denote by $\mathrm{Pred}_v(\mathbf{x}_v)$ the predecessor node of $v$ and its state on the best path to reach $v$'s state $\mathbf{x}_v$. To model that the cumulative number of jumps to reach layer $N$ must be $N-1$ \emph{(b)} at each junction on layer $N>1$, \eqref{eq:setA}~defines $\mathscr{A}$ as the set of admissible combinations of selecting incoming nodes $u_i \in \overleftarrow{N}(v)$ from layers $l_i$. Some admissible sets are visualized in Figure~\ref{fig:suppl:counterex}\textbf{a}.

\section{Applications and Experiments}\label{sec:suppl:app}
\paragraph{Disparity Map Estimation from Stereo Images:}
To generate different disparity maps from stereo images, we build a minimal spanning tree of the pixel grid graph, using the intensity gradient as edge weight as in~\citesuppl{S_Veksler2005}. Neighboring pixels are connected by an edge whenever they have similar intensities. Those that are not similar are not linked and hence are not penalized when generating depth discontinuities. We allow disparities of up to $40$ pixels in either direction while computing matching costs on patches of $11\times 5$ pixels, and use a (non-truncated) quadratic attractive potential on the edges. While this setup is far from state-of-the-art in stereo, it demonstrates that our approach scales to large trees with many labels. We used the proposed diversity accumulation method where one unit of diversity is collected at every state that is at least a distance of $5$ away from the previous solution in label space, and requested a large amount of diversity to obtain visually different depth maps.

\begin{figure}[t!]

\subfigure[]{\includegraphics[width=.32\linewidth]{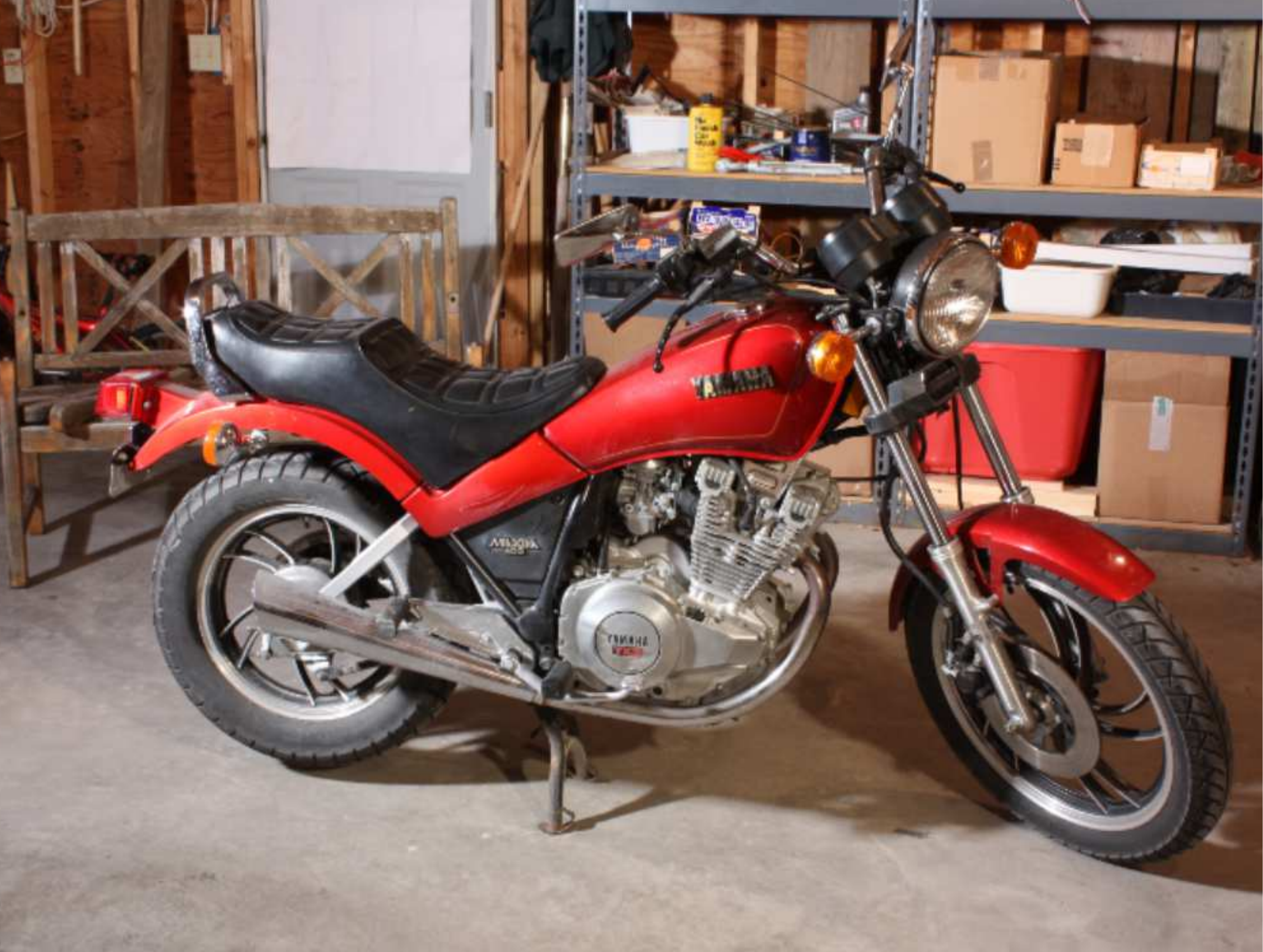}\label{fig:suppl:stereoRaw}}
\subfigure[]{\includegraphics[width=.32\linewidth]{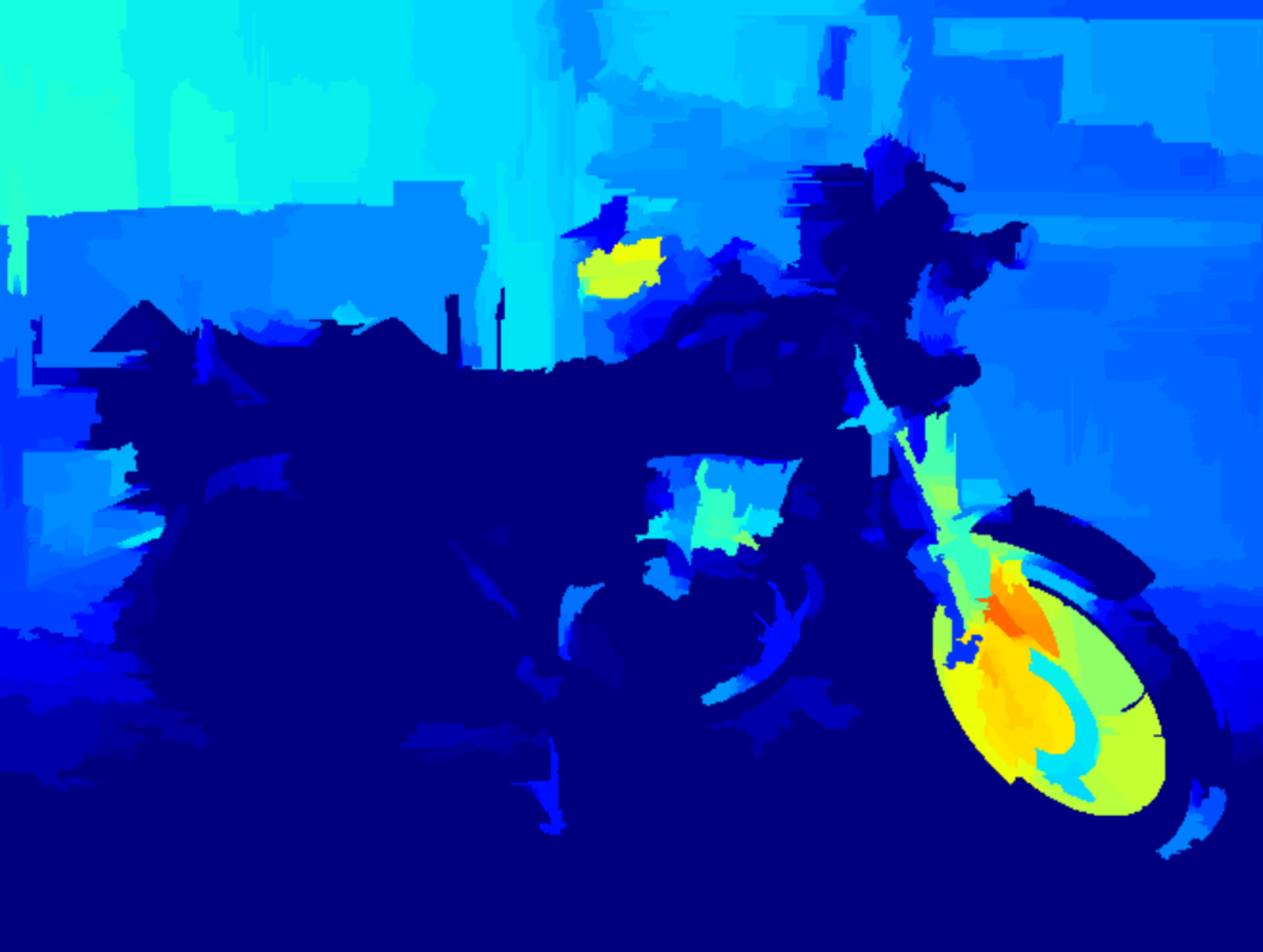}\label{fig:suppl:stereoSol1}}
\subfigure[]{\includegraphics[width=.32\linewidth]{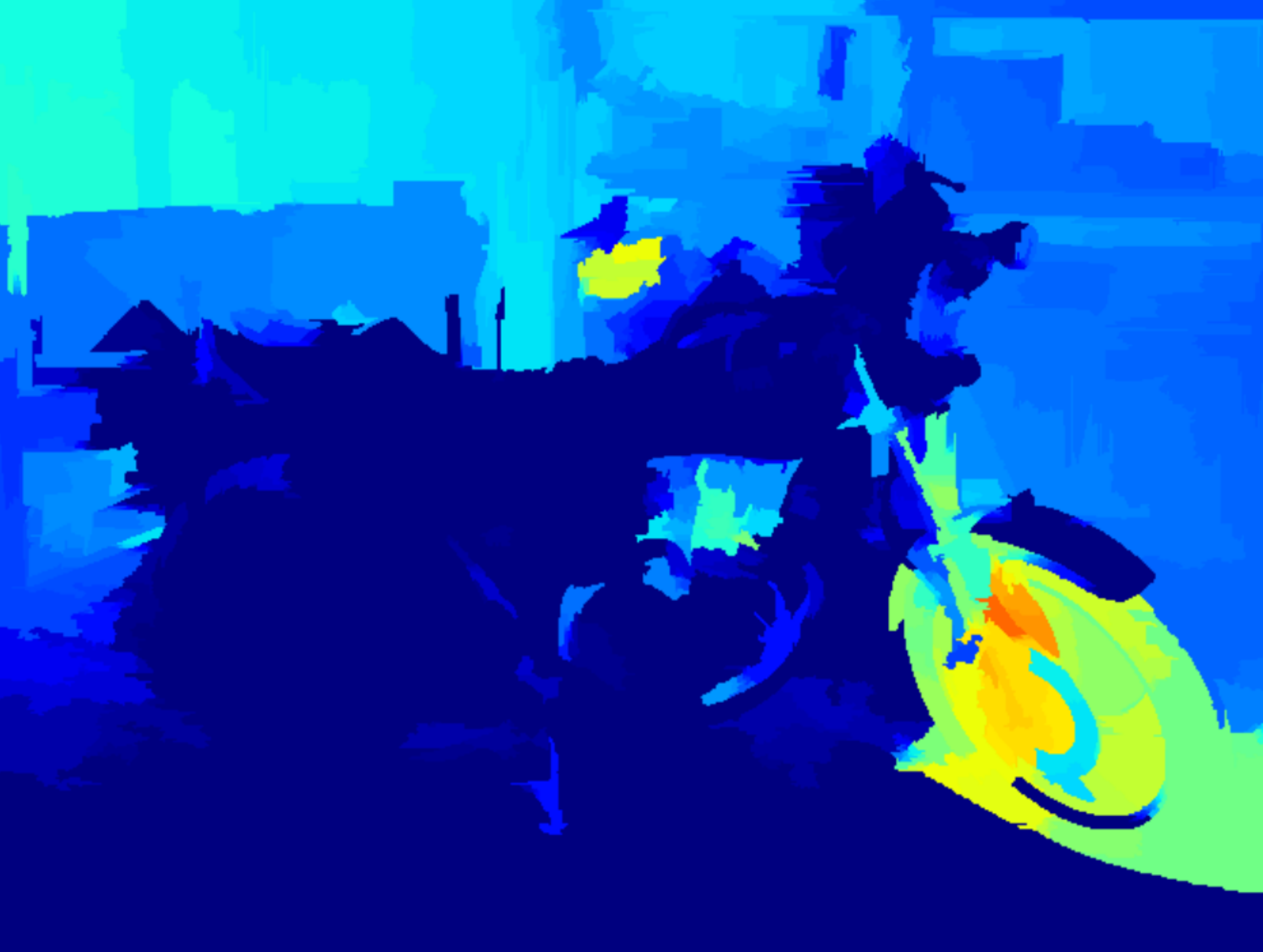}\label{fig:suppl:stereoSol2}}

   \caption{
     Exploring diverse solutions for disparity map estimation, on an image from the Middlebury benchmark~\protect\citesuppl{S_Scharstein2014} resized to $741\times 500$. 
     \textbf{(a)} Left view of the motorbike image pair, and corresponding \textbf{(b)} best solution found by~\protect\citesuppl{S_Veksler2005} which struggles inside the front wheel, probably because the patch size is too large and always contains background and spokes.
     \textbf{(c)} Enforcing a large Hamming distance (here $13000$) reveals that the area around the front wheel could have been matched differently, exposing ambiguities in the estimation process. 
   }
   \label{fig:suppl:stereo}
\end{figure}

\bibliographystylesuppl{splncs03}
\bibliographysuppl{supplementarybiblio}

\pagebreak

\setcounter{equation}{0}
\setcounter{figure}{0}
\setcounter{table}{0}
\setcounter{section}{0}
\renewcommand*{\theHsection}{corrig.\the\value{section}}
\renewcommand*{\theHfigure}{corrig.\the\value{figure}}
\makeatletter
\renewcommand{\theequation}{C\arabic{equation}}
\renewcommand{\thefigure}{C\arabic{figure}}


\title{Corrigendum to the paper:\\Diverse $M$-Best Solutions by Dynamic Programming}

\titlerunning{Corrigendum: Diverse $M$-Best Solutions by Dynamic Programming}
\authorrunning{C. Haubold, V. Uhlmann, M. Unser, F. A. Hamprecht}

\author{Carsten Haubold$^1$, Virginie Uhlmann$^2$, Michael Unser$^2$, Fred A. Hamprecht$^1$}
\date{\vspace{-0.3in}}

\institute{$^1$ University of Heidelberg, IWR/HCI, 69115 Heidelberg, Germany.\\
$^2$ \'Ecole Polytechnique F\'ed\'erale de Lausanne (EPFL), BIG, 1015 Lausanne, Switzerland.}

\maketitle

We\label{chap:corrigendum} have observed that the method presented in section \ref{sec:k1M} of our paper~\citecorrig{Haubold2017Diverse} rested on an assumption that is not always fulfilled.

\begin{figure}
    \centering
    \includegraphics[width=\linewidth]{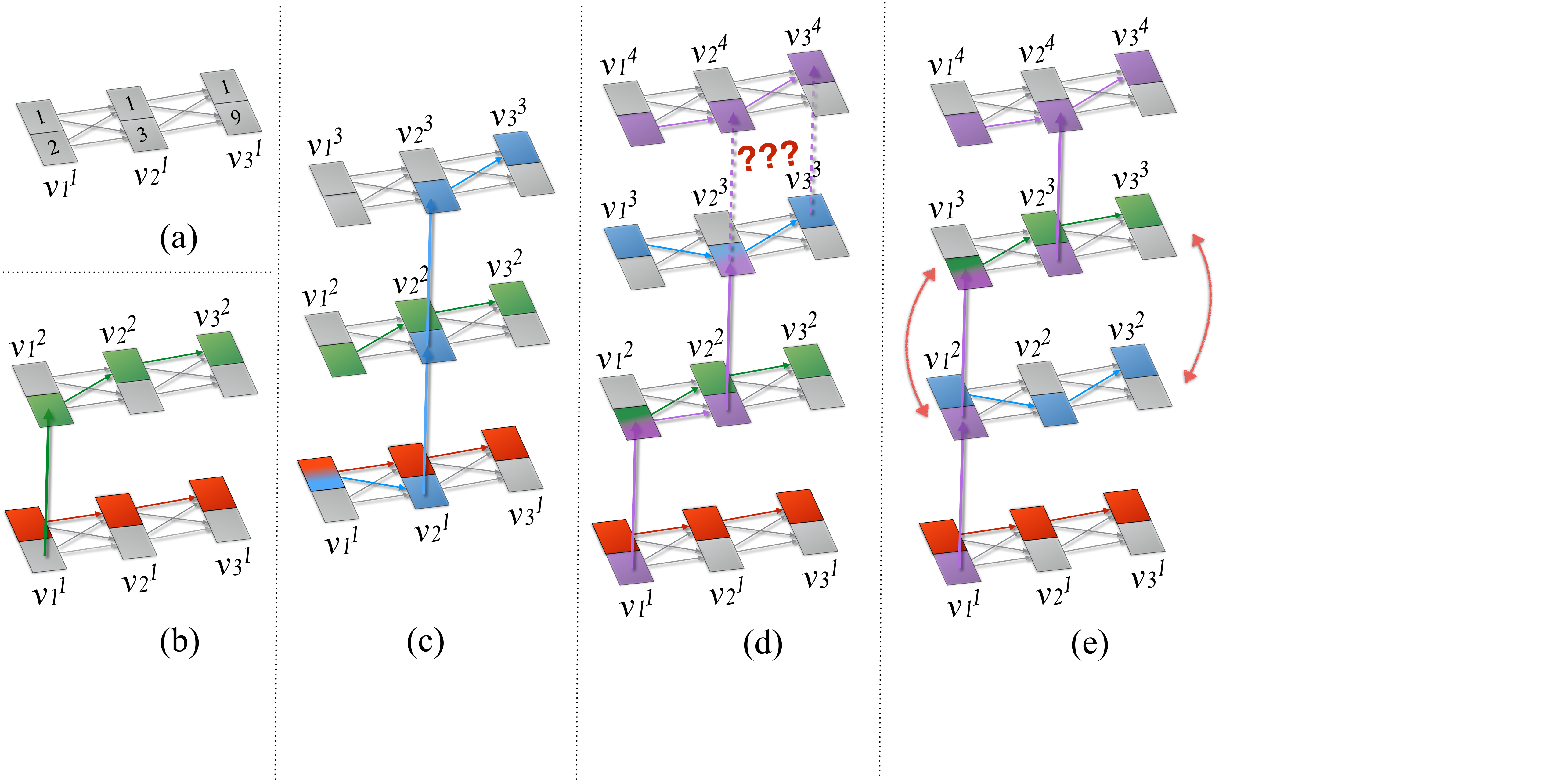}
    \caption{
        \textbf{(a)} Chain model with per-state costs. Assume the edge potentials are uniform. 
        \textbf{(b)} First (red) and second best solution (green) found with 2 layers.
        \textbf{(c)} Third best solution in blue.
        \textbf{(d)} If first and second solution are represented by layers in the order they were found, there is no possibility to jump to the top layer that would give the fourth best solution (purple). Remember that jumps are only allowed at states not used in the solution represented by the layer from which one jumps up.
        \textbf{(e)} When representing the second best solution in the third layer and the third best solution in the second layer, we can find the optimal fourth best solution.
    }
    \label{fig:corrig:counterex}
\end{figure}

Specifically, when looking for the $M^{th}$ best solution in a tree shaped graphical model, when $M>2$ 
the solution found by our method can depend on the order of jumps in previous solutions, and hence does not always represent the optimal $M^{th}$ best configuration.
This is due to the layer ordering from first to $(M-1)^{th}$ best solution. In Figure~\ref{fig:corrig:counterex}, we provide a counterexample indicating that our algorithm sometimes erroneously blocks jumping to the next layer for the next best solution.

A possible way to circumvent this blocking would be to consider every possible ordering of the layers responsible for the $M-1$ previous solutions, find the next best solution in that ordering, and at the end choose the solution that has the minimal cost. This is illustrated in Figure~\ref{fig:corrig:counterex}~\textbf{(e)}.
Unfortunately, this approach increases the runtime complexity by a factor of $(M-1)!$.

The above shows that our method is \emph{not} optimal for the $M^{th}$ best solution with $M > 2$. 
For $M=2$, the scheme we presented in~\citecorrig{Haubold2017Diverse} section \ref{sec:k1M2} remains optimal as a single jump is required to reach the second layer, which can only be blocked by the first best solution. The approximation schemes from section \ref{sec:kM2} were presented for a second best \emph{diverse} solution and are hence not affected. 

\paragraph*{Acknowledgement:} We would like to thank Alexander Richards from Bonn University and the unknown person at GCPR 2017 for their interest and questions that helped us find this error.

\bibliographystylecorrig{splncs03}
\bibliographycorrig{corrigendumbiblio}

\end{document}